\documentclass[11pt,a4paper]{article}
\usepackage[hyperref]{acl2018}
\usepackage{times}
\usepackage{latexsym}
\usepackage{booktabs}
\usepackage{multicol}
\usepackage{epsfig,epsf,subfigure}
\usepackage{siunitx}
\usepackage{url}
\usepackage{xcolor}
\usepackage{todonotes}
\usepackage{makecell}

\aclfinalcopy 

\title{Multimodal Relational Tensor Network for Sentiment and Emotion Classification}

\author{Saurav Sahay\qquad Shachi H Kumar\qquad Rui Xia\qquad Jonathan Huang\qquad Lama Nachman \\
  Anticipatory Computing Lab \\	
  Intel Labs \\
  {\tt \{saurav.sahay, shachi.h.kumar, rui.xia, Jonathan.Huang, lama.nachman\}} \\
      {\tt @intel.com}  \\}

\date{}

\begin{document}
\maketitle
\begin{abstract}
  Understanding Affect from video segments has brought researchers from the language, audio and video domains together. Most of the current multimodal research in this area deals with various techniques to fuse the modalities, and mostly treat the segments of a video independently. Motivated by the work of \cite{ZadehCPCM17} and \cite{Poria2017ContextDependentSA}, we present Relational Tensor Network architecture where we use the inter-modal interactions within a segment and also consider the sequence of segments in a video to model the inter-segment inter-modal interactions. We also generate rich representations of text and audio modalities by leveraging richer audio and linguistic context alongwith fusing fine-grained knowledge based polarity scores from text. We present the results of our model on CMU-MOSEI dataset and show that our model outperforms many baselines and state of the art methods for sentiment classification and emotion recognition.
   
\end{abstract}

\section{Introduction}
Sentiment Analysis is broadly defined as the computational study of subjective elements such as opinions, attitudes, and emotions towards other objects or persons. Sentiments attach to modalities such as text, audio and video at different levels of granularity and are useful in deriving social insights about various entities such as movies, products, persons or organizations. Emotion Understanding is another closely related field that commonly deals with analysis of audio, video, and other sensory signals for getting psychological and behavioral insights about an individual's mental state. Emotions are defined as brief organically synchronized evaluations of major events whereas sentiments on the other hand are considered as more enduring beliefs and dispositions towards objects or persons \cite{scherer1984emotion}. The field of Emotion Understanding has rich literature with many interesting models of understanding \cite{plutchik2001nature} \cite{ekman2009telling} \cite{posner2005circumplex}.

In this work, we explore methods that combine various unimodal techniques for classification alongwith multimodal techniques for fusion of cross modal interactions to perform sentiment analysis and emotion understanding. We develop and test our approaches on the CMU-MOSEI dataset \cite{zadeh2016multimodal} as part of the ACL Multimodal Emotion Recognition grand challenge. CMU Multimodal Opinion Sentiment and Emotion Intensity (CMU-MOSEI) dataset is a newly released large dataset of multimodal sentiment analysis and emotion recognition on YouTube video segments. The dataset contains more than 23,500 sentence utterance videos from more than 1000 online YouTube speakers. The dataset has several interesting properties such as being gender balanced, containing various topics and monologue videos from people with different personality traits. The videos are manually transcribed and properly punctuated. Since the dataset comprises of natural audio-visual opinionated expressions of the speakers, it provides an excellent testbed for research in emotion and sentiment understanding. The videos are cut into continuous segments and the segments are annotated with 7 point scale sentiment labels and 4 point scale emotion categories corresponding to the Eckman'a 6 basic emotion classes \cite{10025007347}. The opinionated expressions in the segments contain visual cues, audio variations in signal as well textual expressions showing various subtle and non-obvious interactions across the modalities for both sentiment and emotion classification.

What differentiates our work from existing literature is (i) application of a novel cross modal fusion technique across the temporal segments of the multimodal channel (ii) use of rich shallow semantic domain knowledge that include a large number of psycholinguistic features and resources for sentiment and emotion classification and (iii) extraction of emotion aware acoustic phoneme level features using a novel method and architecture.

Our unimodal research focus in this paper is an exploration of speech sentiment and emotion recognition using various text dependent and text independent techniques. On the text modality experiments, we've explored (i) fusion of Lexicons as additional input features (ii) fusion of polarity discriminating lexico-syntactic fine-grained scores as additional input features (iii) fusion of rich contextualized embeddings as additional input features to the classification pipeline. On audio modality, we've used a novel pipeline to generate the iVectors and Phoneme level utterance features. For fusion of multimodal information, we have explored techniques that leverage intra-modal and inter-modal dynamics and fused them together in a novel Relational Tensor Network architecture.

\section{Related Work}
Sentiment Analysis has received a lot of prior attention in Movie reviews and Product reviews domain and is an established field of research in NLP \cite{liu2010sentiment} \cite{pang2008opinion}. However, this hasn't been widely researched in conversational multimodal audio-visual and textual context for continuous recognition of sentiments and emotions. \cite{kaushik2013sentiment} perform sentiment extraction on natural audio streams using ASR on Youtube videos. They use a maximum entropy classifier and do not use any lexicon features or do any domain adaptation. Multimodal Affect recognition has lately gained a lot of popularity with release of multiple datasets and approaches \cite{zadeh2018memory}. \cite{ZadehCPCM17} present a tensor fusion technique to generate a fused representation of the individual modalities. Most of these techniques treat the segments of a video independently and ignore the temporal relations and interactions between the segments of a video. \cite{Poria2017ContextDependentSA} present an LSTM based network architecture that leverages the context or the temporal interactions between neighboring segments of a video by concatenation of cross modal features across the segments. For acoustic emotion recognition, one of the most successful system is based on the super-segmental acoustic features which is extracted by applying multiple functions on frame-level features. These features have been adopted as the baseline system in many acoustic emotion challenges \citep{schuller2016interspeech} \citep{valstar2016avec} \citep{dhall2013emotion}. Deep learning techniques have also been used in acoustic emotion recognition system in recent years. In \citep{neumann2017attentive}, convolutional neural network (CNN) is applied on the frame-level feature. In \citep{tao2017advanced}, recurrent neural network (RNN) is used to model the temporal information for emotion recognition system.

\section{Model Description}
This work brings together techniques for various modality specific feature extraction methods and fusion of information from different modalities for Sentiment and Emotion Classification. The grand challenge dataset comes with modality specific features for text, audio and images as a part of the CMU Multimodal Data SDK \cite{zadeh2018multi}. The text features are based on Glove embeddings \cite{pennington2014glove}, audio features are based on COVAREP \cite{covarep}and the visual features based on FACET \cite{7477553} visual feature extraction libraries. We extracted various additional features for text and audio modalities as described in the following sections.

\subsection{Text} 

Several traditional methods have been developed in Sentiment Analysis technology for decades before the recent advances in deep learning that primarily rely on methods for word vector representation and automated feature discovery from snippets. We look at modeling some of the traditional methods and features in the deep pipeline and study the impact of these on the classifiers. Below, we describe a couple of traditional knowledge based resources alongwith some recent deep representations that we have fused together in our pipeline.

\subsubsection{Lexico-syntactic Rule based features}
\begin{figure}
    \includegraphics[scale=0.5]{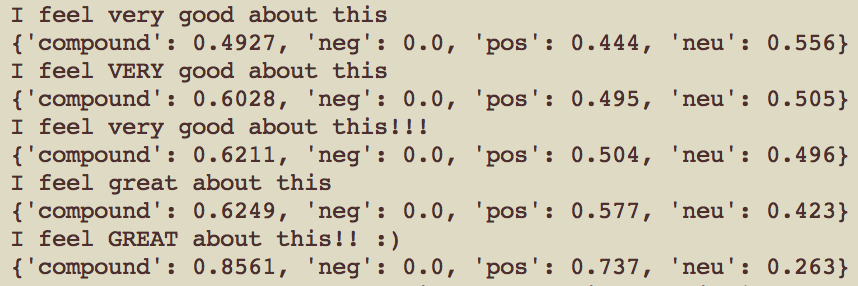}
\caption{Sentiment Analyzer}
    \label{vader}
\end{figure}

Text is processed to intrinsically understand the deeper lexico-syntactic patterns to relate them with world knowledge to extract meaningful inferences such as sentiments and emotions. We have explored the use of VADER rules \cite{HuttoG14} for sentiment and emotion induction. VADER is a simple and fast rule-based model for general sentiment analysis. It utilizes a human-validated general sentiment lexicon and general rules related to grammar and syntax. The goal of this work is to capture generalizable rules and heuristics associated with grammatical and syntactical cues people use to assess sentiment intensity in text. We can clearly see from Figure \ref{vader} how this system can differentiate emphasis, intensity and non-linguistic cues from utterances. Deep learning based systems today fail to capture such systematic nuances deterministically. 

\subsubsection{Sentiment Lexicons}
Lexicons consists of maps of key-value pairs, where the key is a word and the value is a list of sentiment scores for that word (e.g., probabilities of the word in positive, neutral, and negative contexts). The scores have different ranges for showing very negative to very positive sentiments. Lexicon embeddings are sparse signals derived by taking the normalized scores from multiple sources of lexicon datasets. The simplest method of blending a lexicon embedding into its corresponding word embedding is to append it to the end of the word embedding. The General Inquirer(GI) \cite{stone1966general} is a text analysis application with one of the oldest manually constructed lexicons still in widespread use. It contains 11000 words in 183 different psycho-linguistic categories. We have used the lexicon based General Inquirer classes that are divided into groups such as valence, semantic dimensions, cognitive orientation, institutional context, motivation related words, classes of Power, Respect, Affection, Wealth, Well-being, Enlightenment, Skill, etc.. \cite{Shin2017LexiconIC} who originally explored this work in depth show that lexicon embeddings allow building high-performing models with much smaller word embeddings.

\subsubsection{Contextualized Language Embeddings}
In contrast to the above two features, we have also looked at recent developments in contextualized deep word vector representations and how they can help with sentiment and emotion classification. These word vectors are learned functions of the internal states of a deep bidirectional language model, which is pretrained on a large text corpus. The additional language modeling views to vector generation process results in high quality representations \cite{Peters2018DeepCW}. These word vector representations try to model the complex characteristics of word use along with how these uses vary across linguistic contexts (i.e., to model polysemy). We have used ELMo that learns a linear combination of the vectors stacked above each input word for each end task, which improves performance over just using the top LSTM layer  \cite{contextualWV} . Unlike most widely used word embeddings  \cite{pennington2014glove}, ELMo word representations are functions of the entire input sentence

\subsection{Audio Features}
For this task, three different kinds of features were applied. The first one is the feature set
extracted by using COVAREP which is provided by the challenge. It includes multiple kinds of frame level acoustic features,
such as Mel-frequency Cepstral Coefficients (MFCCs), energy and etc. More details are described in \citep{Gusfield:97}.
Along with COVAREP features, we proposed two additional feature-sets, i-vector features and phoneme level features.
Following two sections will discuss details about proposed feature-sets.

\subsubsection{I-vector Features}
The previous studies \citep{xia2016dbn} \citep{tao2018ensemble} have shown that i-vector feature can benefit acoustic emotion recognition system.
I-vector modeling is a technique to map the high dimensional Gaussian Mixture Model (GMM) supervector space (generated by
concatenating the mean of the mixtures from GMM) to low dimensional space called total variability space $T$. 

Give an utterance $u$, $x_t^{u}$ which represents $t$-th frame of utterance $u$.
Audio frame $x_t^{u}$ is generated by the following distribution:
\begin{equation}
\label{eq_ivector}
    x_t^{u} \sim \sum_{c} p(c|x_t^{u})\mathcal{N}( m_{c} + Tw^{u}, \Sigma_{c})
\end{equation}
where $p(c|x_t^{u})$ is the posterior probability of $c$-th Gaussian in Universal Background Model (UBM), $m_{c}$ and $\Sigma_{c}$ represent the means and covariance
of $c$-th Gaussian and $w^{u}$ is the latent i-vector for utterance $u$. 
EM algorithm introduced in is applied to iteratively train $T$.
Note that UBM is a GMM which trained with a large corpus.

In \citep{lei2014novel}, the phonetically-aware DNN is used to replace the traditional UBM in the framework of i-vector training which showed significant improvements on the speaker identification task.
The phonetically-aware DNN is the network for the acoustic model of the Automatic Speech Recognition (ASR) system.
It is trained for recognizing the tri-phone state.
Compared to the traditional trained UBM, the DNN from ASR represents the feature space constrained on pre-defined tri-phone states. 
The posterior probability as the output of this DNN is directly used as the $p(c|x_t^{u})$ in Equation \ref{eq_ivector}.
In this work, ASR and i-vector extractor are pre-trained on Librispeech dataset \citep{panayotov2015librispeech} with Kaldi \citep{povey2011kaldi}.
We used 960 hours speech data from Librispeech to train DNN-HMM ASR and 460 clean data for i-vector extractor.
To avoid overfitting, the dimensionality of i-vector is set as 100. We also tried larger i-vector dimensions but the i-vector with larger dimensions show similar performance compared to the i-vector system with size 100 dimensionality.

\subsubsection{Phoneme Level Features}
The phoneme related information have also been applied in emotion recognition system. 
Phoneme-dependent hidden Markov model (HMM) was proposed for emotion recognition system in \citep{lee2004emotion}.
\citep{bitouk2010class} proposed to extract class-level spectral features on three types of phoneme.
Unlike most other work that need accurate alignment, we propose to use the statistics of posterior probability of phoneme on utterance level.
The following steps are used to extract phoneme level features:

\begin{itemize}
  \item Step One: Each frame $x_t^{u}$ in utterance $u$ is been feed into DNN pre-trained for ASR. 
        The output is a numeric vector consisting of $p(s_i|x_t^{u},DNN)$, which corresponding to posterior probability of triphone state $s_i$. 
        The number of triphone state is dependent on the decision tree algorithm in the ASR system.

  \item Step Two: Mapping the tri-phone state $s_i$ into monophone. The number of the triphone state is huge. 
        For emotion recognition system, it is not necessary to know information in such fine-grained unit. 
        Instead, we map the triphone state into monophone level by disregarding left and right phone in the tri-phone structure.
        The mapping function is: $F_{map}(s_i) = m_j$ where $s_i$ is the tri-phone state and $m_j$ represents corresponding monophone.
        For example, $F_{map}(r-ae-n) = ae$. 

  \item Step Three: Calculating the statistics of posterior probability of phoneme on utterance level.
        Given $x_t^{u}$, for each cluster $m_j$, we sum up the posterior probability $p(s_i|x_t^{u},DNN)$ once $s_i$ belongs to cluster $m_j$.
        \begin{equation}
          P_{m_j}(x_t^{u})  = \sum_{F_{map}(s_i) = m_j} p(s_i|x_t^{u},DNN)
        \end{equation}

        It generates a vector in the length of number of monophone for each frame in utterance $u$.
        In order to obtain a fixed dimensional feature for each utterance with variable length,
        statistics functionals, mean and standard deviation are applied on $P_{M}(X)$.

\end{itemize}
For each utterance, the generated feature set is a fixed dimensional vector.
Based on the trained DNN-HMM ASR system, the number of tri-phone states and monophone ends up in 5672 and 58 respectively.
After mapping and feature extraction, the dimensionality of the phoneme level features is 106.

\section{Network Architectures}

The models described here are based on a recurrent architecture and use different fusion strategies such as concatenation or tensor fusion across all modalities as well as across all segments of the video. We have integrated ideas from the two distinct approaches to jointly leverage multimodal fusion across modalities and across temporal segments and developed our Multimodal Relational Tensor Network.

\subsection{Tensor Fusion Network}

TFN consists of a Tensor Fusion Layer that explicitly models the unimodal, bimodal and trimodal inter-modal interactions using a 3-fold Cartesian product from modality embeddings. Most common deep learning approach for fusion of signals is a algebraic Merge operation where the operators are generally a linear concatenation of features or a sum. TFN, on the other hand, tries to disentangle unimodal, bimodal and trimodal dynamics by modeling each of them explicitly. Tensor Fusion is defined as the three-fold Cartesian product amongst the modalities with an extra constant `1' added to the dimension. The extra constant dimension with value `1' analytically generates all the multimodal dynamics followed with vector dot operations. This definition is mathematically equivalent to a differentiable outer product between the modalities. This operation results is a very large number of dimensions in the merged layer and therefore can realistically be applied to problems where the interaction space is not too large.

\subsection{Contextual LSTM} 
Utterances in a video maintain a continuous sequence and work like state machines following a certain path before changing courses. Statistical Sequence classification techniques are applied in the classification of each member of the sequence by modeling the dependence on the other members of the sequence. Human reactions are also generally continuous and maintain a certain state in the sequence before jumping to another state. In particular, it has been seen that, when classifying one utterance, other utterances can provide important contextual information. This natural phenomenon directly maps to methods such as recurrent network approaches and sequence models to capture the dependencies between the segments. We re-use this idea to capture this flow of informational triggers across utterances using an LSTM-based recurrent neural network (RNN).

\subsection{Relational Tensor Network}

\begin{figure}
\includegraphics[scale=0.4]{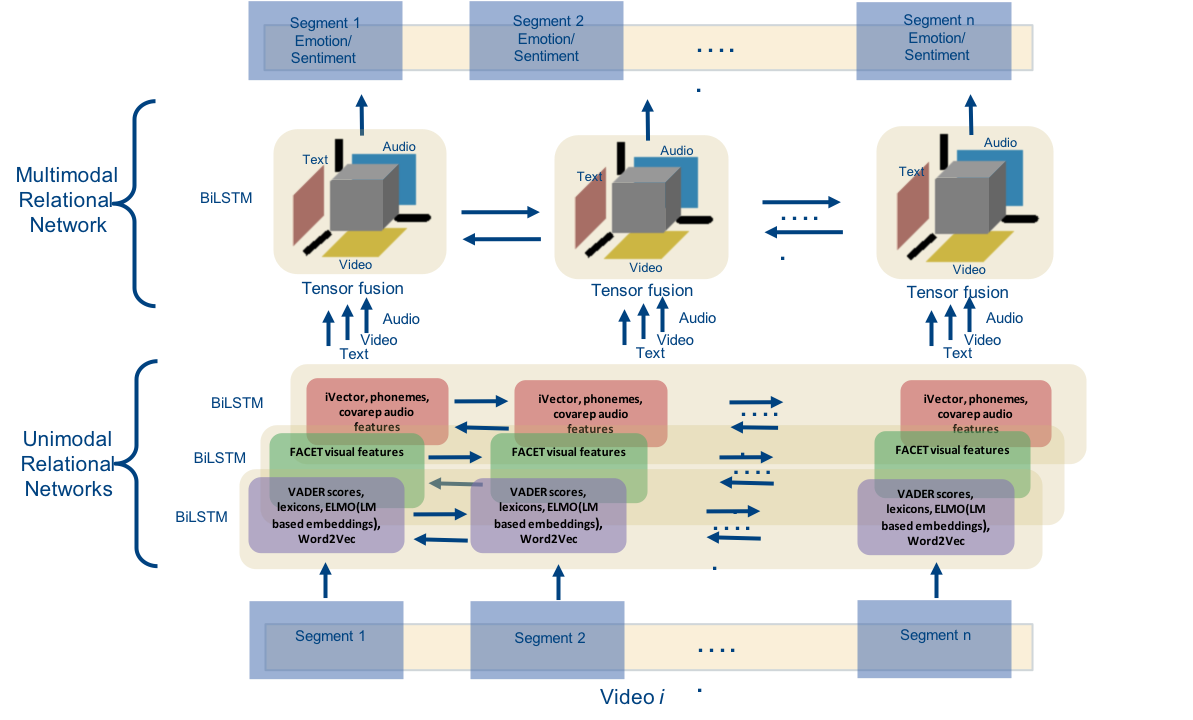}
\caption{Relational Tensor Network}
    \label{arch}
\end{figure}

While TFN has been used to model the modality interactions within a video segment, we extend that approach to apply it to the contextual stream of segments. There are two ways we can apply a tensor fusion (by tensor fusion, we specifically refer to the cartesian product operation between the modalities with an extra `1' input to model the inter-modal interaction) across modalities across the streams. The first approach is to apply a tensor fusion across all modality features of all segments for all the modalities. This approach ideally captures all possible cross-dynamics (unimodal, bimodal, trimodal) amongst all possible features of all the video segments. The main issue with this approach is that we run into an exponential growth in the feature space with every modality added in the interaction. The cartesian product further creates multiple outer products for bimodal and trimodal interactions. Even with a small number of features for this approach, our network had about 10s of billions of parameters and this would not be a feasible approach unless deployed on a massive infrastructure. This approach does not require the use of LSTMs as used in the Contextual LSTM work to capture the sequence information in the segments.
\\
The other more feasible approach is to apply tensor fusion across modality features of each segment and then model the sequential interactions between the segments of the video using an LSTM Network. We depict our network in Figure \ref{arch}.

This approach allows generation of contextually rich features that learn their weights not only from the current rich multimodal interactions but also leveraging previous interactions in the process. For example, interactions amongst audio and text features together can have a multiplicative effect to recognize certain kinds of emotion better (for example, high arousal negative words multiplied together can show stronger bias for the angry emotion). Also, these interactions persist across the segments and can help generate more meaningful recognizer of multimodal interactions. The intuitive explanation of this network is that it captures the long term multiplicative effects of interactions across segments for unimodal, bimodal and trimodal features. Neither the TFN model or the contextual model alone can effectively capture these interactions in principle.

\section{Experiments}
We present multiple sets of experiments in order to evaluate the different models, impact of textual and audio features on sentiment and emotion prediction. Our training data consists of CMU-MOSEI training set where we do a 90/10 split for validation and early stopping experiments. All our results in this paper are reported on the CMU-MOSEI validation set\footnote{\footnotesize{The test set was not released at the time of writing.}}. 

\begin{table}
\begin{tabular}{|S|SS|SS|S|} \toprule
\multicolumn{1}{|c|}{} & \multicolumn{2}{c|}{\scriptsize Binary}	 & \multicolumn{2}{c|}{\scriptsize 7-class} &  \multicolumn{1}{c|}{\scriptsize Regression} \\
\cline{2-3}  \cline{4-4} \cline{5-6}
                {\scriptsize Baseline} & {\scriptsize Acc} & {\scriptsize F1} & {\scriptsize Acc} & {\scriptsize F1} & \scriptsize {MAE}  \\ \midrule
      
      \scriptsize {$ {SVM \; multimodal}$}    &\scriptsize 60.4&\scriptsize  0.61   &\scriptsize 23.5 &\scriptsize 0.27     &\scriptsize 1.38 \\ 
     \scriptsize {$LSTM\_uni\_audio$}     &\scriptsize 58 & \scriptsize 0.52   &\scriptsize 41 &\scriptsize 0.37     &\scriptsize 0.73 \\
       \scriptsize {$LSTM\_uni\_video$}     &\scriptsize 57.9 & \scriptsize 0.51 &\scriptsize  45.9 &\scriptsize  0.40&\scriptsize  0.68\\
       \scriptsize {$LSTM\_uni\_text$}     &\scriptsize 64.2 & \scriptsize 0.60   &\scriptsize 45.8 &\scriptsize 0.43     &\scriptsize 0.618 \\
       \scriptsize {$LSTM\_early fusion$}     &\scriptsize 65.2& \scriptsize 0.62  &\scriptsize 46.6 &\scriptsize 0.44 &\scriptsize 0.60  \\       
\scriptsize{$\;\;\;\;\;\;\;\;\;\;\;\;TFN$}    &\scriptsize 66&\scriptsize 0.62   & \scriptsize 47.9 &\scriptsize 0.43     &\scriptsize 0.58 \\     

           \scriptsize{$\;\;\;\;\;\;\;\;\;\;\;\;RTN$}    &\scriptsize \textbf{66.8 }&\scriptsize  \textbf{0.63}   &\scriptsize \textbf{49.17} &\scriptsize \textbf{0.45}     &\scriptsize \textbf{0.58} \\
      
      \bottomrule      

    \end{tabular}
    \caption{Sentiment Analysis Model Results}
    \label{model_results}
\end{table}

 \begin{table*}
\centering
\begin{tabular}{SSSSSSS} \toprule
& {\scriptsize $Anger$}& {\scriptsize $Disgust$} &  {\scriptsize $Fear$} &  {\scriptsize $Happy$}&  {\scriptsize $Sad$} &  {\scriptsize $Surprise$} \\ \toprule
{\scriptsize $ {SVM \; multimodal}$}& \scriptsize 0.358 & \scriptsize 0.19 &\scriptsize 0.21 &\scriptsize 1.167 &\scriptsize 0.33 &\scriptsize 0.171  \\
{\scriptsize $LSTM\_uni\_audio$} &\scriptsize 0.17 &\scriptsize 0.079 &\scriptsize 0.09 &\scriptsize 0.475 &\scriptsize 0.20 &\scriptsize 0.073  \\
{\scriptsize $LSTM\_uni\_text$}&\scriptsize 0.16&\scriptsize 0.08&\scriptsize 0.086&\scriptsize 0.485&\scriptsize 0.195&\scriptsize 0.068 \\
{\scriptsize $LSTM\_uni\_video$} &\scriptsize 0.148&\scriptsize 0.08&\scriptsize 0.10 &\scriptsize 0.42&\scriptsize 0.208&\scriptsize 0.076 \\
{\scriptsize $LSTM\_early fusion$} &\scriptsize 0.148 &\scriptsize 0.078 &\scriptsize 0.09 &\scriptsize 0.428 &\scriptsize 0.19 &\scriptsize 0.073  \\
\scriptsize {$\;\;\;\;\;\;\;\;\;\;\;\;TFN$} &\scriptsize 0.147&\scriptsize 0.07&\scriptsize 0.089&\scriptsize 0.466&\scriptsize 0.1766 &\scriptsize 0.074  \\
\scriptsize {$\;\;\;\;\;\;\;\;\;\;\;\;RTN$} &\scriptsize \textbf{0.137} &\scriptsize \textbf{0.065} &\scriptsize \textbf{0.072} &\scriptsize \textbf{0.422} &\scriptsize \textbf{0.176} &\scriptsize \textbf{0.059} \\ \bottomrule
    \end{tabular}
       \caption{Emotion Recognition Model Results - MAE scores}
    \label{model_results_emotion}
\end{table*}

 \begin{table*}
\centering
\begin{tabular}{SSSSSS} \toprule
\multicolumn{1}{c}{} & \multicolumn{2}{c}{\footnotesize Binary} & \multicolumn{2}{c}{\footnotesize 7-class} &  \multicolumn{1}{c}{\footnotesize Regression} \\
\cline{2-3}  \cline{4-4} \cline{5-6}
           \scriptsize     {Baseline} &\scriptsize {Acc} &\scriptsize {F1} &\scriptsize {Acc} &\scriptsize {F1} &\scriptsize {MAE}  \\ \midrule

              \scriptsize       {$Embedding \; only$}    &\scriptsize 64.6 &\scriptsize 0.60  &\scriptsize 48.17 &\scriptsize 0.43    &\scriptsize 0.595 \\ \midrule
              \scriptsize         {$Emb + Lex$}    &\scriptsize 62.8&\scriptsize   0.57   &\scriptsize 45.6 &\scriptsize 0.41     &\scriptsize 0.61\\
\scriptsize    {$Emb+Vader$}    &\scriptsize 64.2&\scriptsize  0.59   &\scriptsize 45.4 &\scriptsize 0.42     &\scriptsize 0.61 \\
\scriptsize {$Emb+ELMO$}    &\scriptsize 65.5 &\scriptsize  0.61   &\scriptsize 47.5  &\scriptsize  0.44     &\scriptsize 0.589 \\
 \midrule
 
    
 \scriptsize        {$Emb+Lex+Vader$}  &\scriptsize 64.2 &\scriptsize 0.59  &\scriptsize 47.5 &\scriptsize 0.44&\scriptsize 0.59   \\
\scriptsize     {$Emb+ELMO+Lex$}    &\scriptsize 64.6   &\scriptsize{0.58}&\scriptsize {48.7} &\scriptsize {0.45}    &\scriptsize 0.576 \\
\scriptsize         {$Emb+ELMO+Vader$}  &\scriptsize \textbf{{66.4}} &\scriptsize \textbf{{0.63}}&\scriptsize \textbf{{48.9} } & \scriptsize \textbf{{0.44}}&\scriptsize \textbf{0.577} \\
\midrule
        
\scriptsize {$ All \; features$}    &\scriptsize 66&\scriptsize 0.62   &\scriptsize 47.9 &\scriptsize  0.43     &\scriptsize 0.58 \\ \bottomrule
    \end{tabular}
\caption{Text Ablation Study}
\label{ablation_text}
\end{table*}

\subsection{Architecture comparisons}
Table \ref{model_results} and Table \ref{model_results_emotion} show the performance of the various models on sentiment and emotion classification. We have used three LSTM based unimodal baselines, each for audio, video and text modalities. From the table, we see that unimodal-text network outperforms both audio and video modalities for sentiment. Unimodal-text also outperforms SVM multimodal for sentiment analysis, which is an SVM model trained on concatenated features from all the three modalities. The early fusion network is an LSTM based network(an extension of the unimodal networks),that takes in concatenated features from the three modalities. This LSTM model outperforms the SVM multimodal baseline by almost 5\% binary class accuracy scores for sentiment analysis. All of these LSTM based networks outperform SVM by a huge margin in the 7-class classification scores and MAE for sentiment analysis.  The $TFN$ network with rich set of textual features slightly outperforms the simple concatenation technique(early fusion model) for sentiment and emotion recognition. The model with the best performance is the Relational Tensor Network model for both sentiment and emotion recognition that considers the neighboring tensor fusion networks for a given segment.

\subsection{Ablation study}
 Table \ref{ablation_text} shows the detailed ablation study of the various text features that we have used in our models. We added word based features using lexicons and language model based ELMo embeddings and utterance level sentiment scores using VADER scores. As the table shows, adding lexicons result in a slight drop in performance of the scores. The lexicons we've used are extremely sparse compared to the vocabulary space of Word Vectors. Also we've simplistically concatenated the binary scores for Positive and and Negative category words to the same embeddings space as for the word vectors.  Majority of these values remain 0 after the operation. We are exploring other ways to leverage the lexicon embeddings to allow a larger contribution of these signals to the classification process. Addition of the ELMo embeddings  improves the performance as compared to using word embeddings alone. Addition of ELMo embeddings and segment level sentiment scores using Vader gives the best performance for binary, 7-class and MAE scores, as compared to adding individual features, or a combination of features. As described in ELMo work, adding the layers at different positions of the network helps to abstract various naturally occurring syntactic and semantic information about the words. \
For the audio modality, we presented two additional feature-sets in the previous section, i-vector features and phoneme level features alongwith COVAREP features. Based on our experiments, we observed that the performance of the Emotion recognition RTN model with all these features were similar but improved slightly for `Happy' emotion compared to the RTN model without the additional audio features.

%

\section{Conclusion}
In this paper we present a novel model called Relational Tensor Network for multimodal Affect Recognition that takes into account the context of a segment in a video based on the relations and interactions with its neighboring segments within the video. We meticulously add various feature set on the word level, that involves language model based embeddings and segment level sentiment features. Our model shows the best performance as compared to the state of the art techniques for sentiment and emotion recognition on the CMU-MOSEI dataset.\\

\bibliography{acl2018}

\bibliographystyle{acl_natbib}

\end{document}